\documentclass{ifacconf}

\makeatletter
\let\old@ssect\@ssect 
\makeatother

\usepackage{graphicx}      
\usepackage{natbib}        

\usepackage{mydefinitions} 
\usepackage{irtdefinitions} 
\usepackage{etoolbox}

\makeatletter
\def\@ssect#1#2#3#4#5#6{%
  \NR@gettitle{#6}
  \old@ssect{#1}{#2}{#3}{#4}{#5}{#6}
}
\makeatother
\begin{document}
\begin{frontmatter}

  \title{Informed Circular Fields for Global Reactive Obstacle Avoidance of Robotic Manipulators\thanksref{footnoteinfo}}

  \thanks[footnoteinfo]{This work was supported in part by the Region Hannover in the project \emph{roboterfabrik}.}

  \author[First]{Marvin Becker}
  \author[First]{Philipp Caspers}
  \author[First]{Tom Hattendorf}
  \author[First]{Torsten Lilge}
  \author[Second]{Sami Haddadin}
  \author[First]{Matthias A. M{\"u}ller}

  \address[First]{Institut of Automatic Control, Leibniz University Hannover, Germany (e-mail: \{becker,lilge,mueller\}@irt.uni-hannover.de)}
  \address[Second]{ Munich School of Robotics and Machine Intelligence and Chair of Robotics and System Intelligence, Technical University Munich, Germany (e-mail: sami.haddadin@tum.de).}

  \begin{abstract}                
    In this paper a global reactive motion planning framework for robotic manipulators in complex dynamic environments is presented.
    In particular, the circular field predictions (CFP) planner from \cite{BeckerLilMulHad2021} is extended to ensure obstacle avoidance of the whole structure of a robotic manipulator.
    Towards this end, a motion planning framework is developed that leverages global information about promising avoidance directions from arbitrary configuration space motion planners, resulting in improved global trajectories while reactively avoiding dynamic obstacles and decreasing the required computational power.
    The resulting motion planning framework is tested in multiple simulations with complex and dynamic obstacles and demonstrates great potential compared to existing motion planning approaches.
  \end{abstract}


\end{frontmatter}
%
%
\section{Introduction}

\vspace{-1ex}
\emph{Motivation:}
Classical industrial robotic applications require fences or other peripheral safety installations to ensure seamless production processes and the safety of human coworkers.
However, these measures result in increasing costs and larger space requirements while decreasing flexibility.
As a consequence, robots, especially collaborative systems, are increasingly exposed to dynamic and unpredictable environments requiring collision-free motion planning.
However, collision avoidance of robotic manipulators in dynamic environments remains a major challenge \citep{LiuQuXuDu2022}.\\
\emph{Related Work:}
Due to the amount of research on motion planning, we focus on sampling-based and reactive approaches, using the classification of motion planners into \emph{sense-plan-act}, \emph{locally reactive control} and \emph{reactive planning} \citep{KapplerMeiIssMai2018}.
Traditional approaches typically use the \emph{sense-plan-act} paradigm with a rather strict separation of perception, motion planning and control. Most sampling-based approaches belong to this category.
Among those, the \gls{rrt} algorithm \citep{LaValle1998} and the \gls{prm} approach \citep{KavrakiSveLatOve1996} are the most widely used concepts.
The \gls{rrt} algorithm incrementally constructs a search tree with randomly selected nodes in the robot's configuration space to find a feasible path from the initial to the goal pose.
Due to the ease of implementation, many successful applications, and the simple extension to higher dimensional problems, the \gls{rrt} planner has been improved and extended to different variants, e.g., the \gls{rrt}-connect improves the runtime \citep{KuffnerLaV2000}, whereas the asymptotically optimal \gls{rrt}* decreases the resulting path length \citep{KaramanFra2011}.
The \gls{prm} algorithm consists of a learning and a query phase. During the learning phase, the robot's configuration space is randomly sampled and configurations that collide with obstacles are rejected.
The resulting roadmap is then used for the query phase, which calculates the shortest path from a start to a goal pose.
Extensions of the \gls{prm} planner focus on improved path quality (\gls{prm}* \citep{KaramanFra2011}) or shorter runtimes (Lazy\gls{prm}* \citep{Hauser2015}).
Although the improvements of sampling based planners significantly decrease computation costs and show good results even in dynamic environments, \cite{KapplerMeiIssMai2018} has shown the advantages of reactive approaches over the classical sense-plan-act methodology, particularly in highly dynamic or uncertain environments.\\
While sampling-based approaches generally perform global planning, locally reactive control methods only consider obstacles in direct proximity and calculate the immediate next control command, allowing instantaneous reactions to environment changes.
A widespread example is the \gls{apf} approach \citep{Khatib1986}, where the robot is controlled by artificial repulsive and attractive forces for a collision-free motion to the goal pose.
Other approaches use repulsive forces or velocities for collision avoidance of the whole structure of the robot, which were either applied on predefined control points along the robot structure \citep{ChenSon2018} or use the closest distance between robot and obstacle \citep{MaciejewskiKle1985}.
Another alternative is the \gls{cf} approach \citep{SinghSteWen1996} generating artificial forces similar to forces on charged particles in electromagnetic fields. The approach gained recent interest due to its ability to smoothly guide the robot around obstacles without local minima \citep{HaddadinBelAlb2011,AtakaLamAlt2018,LahaFigVraSwi2021,LahaVorFigQu2021}.
However, immediate reactions and local sensor information prevent locally reactive approaches from finding optimal global paths.\\
\cite{KapplerMeiIssMai2018} introduce the term \emph{reactive planner} for hybrid motion planners that are able to quickly react to local changes while simultaneously improving the global path in case of larger changes.
Hybrid approaches, generally using a combination of \gls{prm} or \gls{rrt} and \gls{apf}, have been mainly applied in mobile robotics, e.g., \citep{RavankarRavEmaKob2020}, but also for manipulators \citep{LiHanLiZha2021}.
Other hybrid approaches are the popular elastic strip framework \citep{BrockKha2002}, or our recent \gls{cfp} planner \citep{BeckerLilMulHad2021}, which extends the \gls{cf} approach with a predictive multi-agent framework for global path exploration.
The \gls{cfp} planner showed promising results in terms of computation load and path quality and was successfully applied on a 7-\gls{dof} manipulator.
For this planner, collision avoidance and goal convergence under defined conditions were rigorously proven in \cite{BeckerKoeHadMul2022}.
Nevertheless, the approach was so far only used to achieve obstacle avoidance of the robot's \gls{ee}.\\
\emph{Contribution:}
The contributions of this paper include:
\begin{itemize}\vspace{-1ex}
  \item Extension of the \gls{cfp} planner for full body obstacle avoidance of robotic manipulators by introducing additional control points along the robot structure and defining suitable control forces.
  \item Integration of global environment information from arbitrary global planners in the multi-agent framework from \cite{BeckerLilMulHad2021} resulting in the \gls{icf} planner.
  \item Two algorithms for leveraging global information about promising avoidance directions with the reactive \gls{icf} planner.
  \item Extensive comparison of the \gls{icf} planner against widely used global and local motion planning approaches in a total of more than 200 simulations.
\end{itemize}
%
%
\section{Full Body Avoidance using Circular Field Predictions}\label{sec:cfp_full_body}

\vspace{-1ex}
In this section, we introduce the natural extension of the \gls{cfp} planner from \cite{BeckerLilMulHad2021} for full body avoidance of a robotic manipulator. We use a steering force for controlling the \gls{ee}
\begin{equation}\label{eq:steering_force_ee}
  \fsee = \fcf + \kvlc \fvlc,
\end{equation}
that consists of the \gls{cf} force $\fcf$ (\cref{sec:ee_obstacle_force}) for avoiding obstacles and an attractive potential force $\fvlc$ (\cref{sec:goal_force}) for goal convergence. The scaling factor $\kvlc \geq 0$ of $\fvlc$ is explained in detail in \cref{sec:goal_force}. While $\fcf$ only contains translational forces, $\fvlc$ additionally includes rotational components to reach the desired goal pose.
%
\subsection{Attractive Goal Force}\label{sec:goal_force}

\vspace{-1ex}
As in \cite{BeckerLilMulHad2021} we use the \gls{vlc} from \cite{Khatib1986} instead of a simple attractive potential. This ensures that the robot moves at a constant maximum velocity during the nominal motion.
Additionally, we define the scaling factor $\kvlc$ as
\begin{equation}\label{eq:k_vlc}
  \kvlc = \begin{cases}
    0 & \! \text{if }\dot{\xv} \!\cdot\!\! \fvlc \!\leq 0 \land \norm{\dot{\xv}} \!\leq\! v_{\min} \land \norm{\xgoal - \xv} \!>\! \xi \\
    w & \! \text{otherwise}
  \end{cases}
\end{equation}
with the scaling factor $w = w_1 w_2 w_3$ and $w_1, w_2, w_3 \geq 0$ from \cite{AtakaLamAlt2018}, which are used to prioritize obstacle avoidance over goal convergence.
The \gls{cf} force does not change the magnitude of the robot velocity, which is therefore only modified by the \gls{vlc}. Using \cref{eq:k_vlc} and therefore deactivating $\fvlc$ when it works against the current motion direction while the velocity is below or equal to a defined $v_{\min}$, we ensure that the robot will only decrease its velocity below this minimum, when the robot is in the vicinity $\xi >0$ of the goal pose.
%
\subsection{Endeffector Obstacle Avoidance}\label{sec:ee_obstacle_force}

\vspace{-1ex}
We use a similar \gls{cf} force definition as \cite{BeckerLilMulHad2021} for the obstacle avoidance of the \gls{ee} and
assume that obstacle data is received in point cloud format, which is commonly used by robotic sensors like depth cameras or laser scanners.
The \gls{cf} force for a single point on a point cloud obstacle is defined as
\begin{equation}
  \fcf := \frac{\kcf}{\norm{\dv}}\, \ubar{\dot{\dv}} \times \Bm
\end{equation}
with the notation $\ubar{\dot{\dv}} = \frac{\dot{\dv}}{\norm{\dot{\dv}}}$ and
the constant gain $\kcf \geq0$, the distance $\dv=\xv_\mathrm{o} - \xv$ and the relative translational velocity $\dot{\dv}$ between the obstacle at position $\xv_\mathrm{o}$ and the robot.
The artificial magnetic field $\Bm$ is defined as $\Bm := \cvec \times \ubar{\dot{\dv}}$ with the artificial electric current
$\cvec := \nv \times \bv$, where $\nv$ is the normalized surface normal of the point cloud point\footnote{Note that many approaches for the surface normal approximation of point clouds exist in the literature. We use the provided functionality of the Point Cloud Library \citep{RusuCou2011}.} and $\bv$ is the artificial magnetic field vector, which is used to define the avoidance direction around an obstacle. The artificial magnetic field vector is therefore of crucial importance for the avoidance maneuver and different possibilities for a meaningful choice are introduced in \cref{sec:multi_agents,sec:global_mag_field}.
The total avoidance force is the sum of the forces of all points $m_j$ on all obstacles $n_\mathrm{o}$
\begin{equation}\label{eq:cf_force}
  \fcf = \frac{1}{\sum^{n_\mathrm{o}}_{j=0}m_j}\sum^{n_\mathrm{o}}_{j=0} \sum^{m_j}_{k=0} \fv_{\mathrm{cf}_{j,k}}.
\end{equation}
In order to save computational resources, only obstacle points $k$ on surfaces facing the robot, i.e. $\nv_k \cdot \dv_k < 0$, and in a range $\norm{\dv_k} \leq \drange$ around the robot are used.
However, in contrast to a majority of approaches in the literature, we clearly exploit more information about the obstacles than only the obstacle point with the minimum distance.
This makes our algorithm less sensitive to sensor noise and yields improved avoidance behavior in the presence of multiple obstacles without oscillations.
%
\subsection{Robot Body Obstacle Avoidance}\label{sec:body_obstacle_force}

\begin{figure}
  \centering
  \includegraphics[trim={11cm 7cm 17cm 4cm},clip,width=0.4\columnwidth]{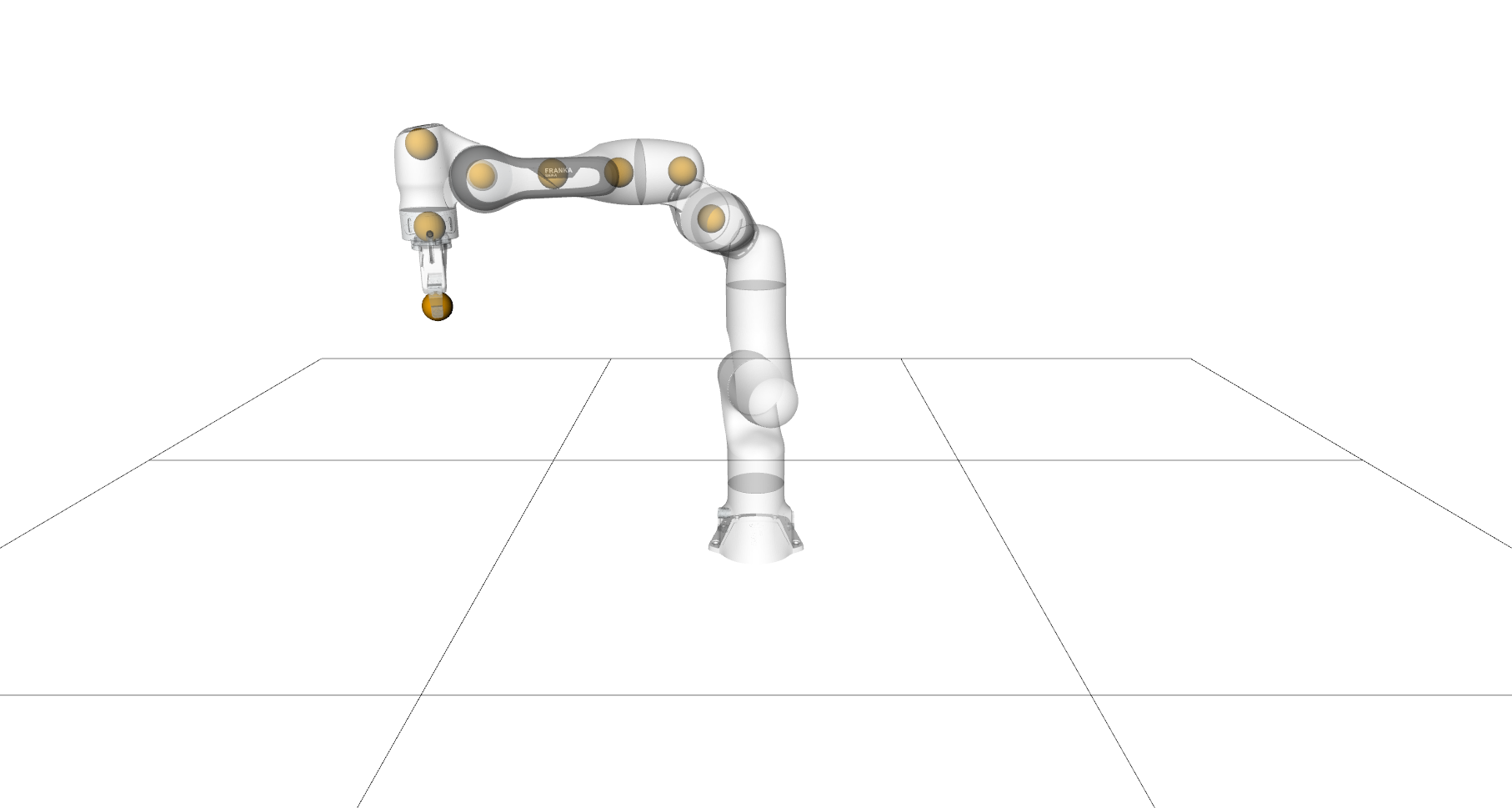}
  \vspace{-1ex}
  \caption{Visualization of possible robot control points.}
  \label{fig:control_points}
\end{figure}

\vspace{-1ex}
To enable obstacle avoidance for the entire robot body, we define $n_\mathrm{cp}$ additional control points along its structure, as exemplarily shown for a 7-\gls{dof} robot arm in \cref{fig:control_points}.
The control points should be placed on prominent points of the robot so that the whole structure can be moved away from obstacles.
Note that for the example in \cref{fig:control_points}, we have refrained from adding more control points on the lower links as their obstacle avoidance capabilities are limited.
Those additional control points are also subject to the \gls{cf} avoidance forces to guide them around obstacles.
However, the motion of these control points may be constrained and the guiding force in \cref{eq:cf_force} might not be sufficient to keep a safe distance to the obstacle.
Towards this end, we add an additional repulsive \gls{cf}-like force, where the artificial current is defined as the negative distance vector
\begin{equation}\label{eq:cf_repulsive}
  \fr := \ubar{\dot{\dv}} \times \left(-\ubar{\dv} \times \ubar{\dot{\dv}} \right) f(\norm{\dv}),
\end{equation}
where we use the logistic amplitude function $f(\norm{\dv})=\frac{1}{2}(1 + \tanh(\alpha - \beta \cdot \norm{\dv}))$ from \cite{LuoKoChuCha2014} with the tunable parameters $\alpha, \beta \geq 0$.
Note that \cref{eq:cf_repulsive} is a simplification of the force from \cite{AtakaLamAlt2022}, which pushes the control point away from the obstacle point while maintaining the useful properties of the original \gls{cf} force, i.e., it is perpendicular to the robot velocity and therefore does not induce local minima.
Subsequently, the total avoidance force on a control point is $\fcp = \fcf + \fr$.
%
\subsection{Robot Control Signal}\label{sec:robot_control}

\vspace{-1ex}
In this section we describe how we generate feasible desired joint velocities for a robotic manipulator from the task space avoidance forces.
Towards this end, we consider the \gls{ee} motion to be the primary task and first calculate a desired joint velocity from the current \gls{ee} velocity $ \dot{\qv}_{\mathrm{d}} = \Jm_{\mathrm{ee}}^\#(\qv) \dot{\xv}_{\mathrm{ee}}$ using the Moore-Penrose pseudo inverse $\Jm_{\mathrm{ee}}^\#$.
The artificial steering force is interpreted as a force on a unit mass, i.e., $\ddot{\xv}_\mathrm{ee} = \fsee$. Then, the operational space formulation from \cite{Khatib1987} is used to calculate a meaningful robot input
\begin{equation}\label{eq:khatib_cartesian}
  \tauv_\mathrm{ee} = \Jm_{\mathrm{ee}}^T(\qv) \left( \Mm_{\mathrm{c}}(\qv) \fsee + \cvec_{\mathrm{c}}(\qv, \dot{\qv}) + \gv(\qv)  \right),
\end{equation}
with the Cartesian mass matrix $\Mm_{\mathrm{c}}(\qv)$, the Coriolis and centrifugal terms $\cvec_{\mathrm{c}}(\qv, \dot{\qv})$, and the gravitational forces $\gv(\qv)$.
Note that we do not consider real forces but want to transform the artificial task space forces into desired joint velocities as inputs of an appropriate joint velocity controller that compensates gravity and dynamics.
Therefore, we ignore gravitational and Coriolis terms and set the mass matrix to the identity matrix $\Imat$, which implies $\ddot{\qv}_\mathrm{ee}=\tauv_\mathrm{ee}$ and leads to $\Mm_{\mathrm{c}}(\qv) = \left(\Jm_{\mathrm{ee}}(\qv) \Imat \Jm_{\mathrm{ee}}^T(\qv)\right)^{-1}$.
Inserting $\Mm_{\mathrm{c}}(\qv)$ into \cref{eq:khatib_cartesian} and neglecting the gravitation and dynamic components yields $\ddot{\qv}_\mathrm{ee} = \Jm_{\mathrm{ee}}^\#(\qv) \fsee$.
The forces on the control points do not need to follow any given dynamics and are superposed and converted to joint accelerations conventionally:
\begin{equation}
  \ddot{\qv}_{\mathrm{cp}} = \sum_{i=0}^{n_\mathrm{cp}-1} \Jm_{\mathrm{cp},i}^T(\qv) \, \fv_{\mathrm{cp},i}.
\end{equation}
Finally, we calculate the joint velocity command for the next sampling step of the controller
\begin{equation}\label{eq:q_cmd}
  \dot{\qv}_\mathrm{cmd} = \dot{\qv}_{\mathrm{d}} + \left( \ddot{\qv}_{\mathrm{ee}} + \ddot{\qv}_{\mathrm{cp}} + \ddot{\qv}_{\mathrm{jla}} \right) \Delta T,
\end{equation}
with the sampling time $\Delta T$. Note that we also added a simple spring-like acceleration $\ddot{\qv}_{\mathrm{jla}}$ for avoiding joint limits.
%
\subsection{Predictive Multi-Agent Framework}\label{sec:multi_agents}
\vspace{-1ex}
In order to transform the original locally reactive \gls{cf} approach into a reactive planner, we extend the predictive multi-agent framework from \cite{BeckerLilMulHad2021} to robotic manipulators.
The framework is used to evaluate the influence of selected control parameters on the robot in a current perception of the environment under simplified dynamics.
A predictive agent represents the robot in this environment snapshot with one specific set of parameters $\mathbb{P}$.
In this paper we use the multi-agent framework primarily for the magnetic field vector $\bv \in \mathbb{B} \subset \mathbb{R}^{n_\mathrm{cp}\times n_\mathrm{o}}$ of the  $n_\mathrm{cp}$ control points (including the \gls{ee}).
After an initial agent with the current robot configuration is created, the joint velocity commands are calculated as described in \cref{eq:q_cmd}.
Then, we simulate the motion of the robot, assuming simplified dynamics and that the controller follows the joint commands perfectly, i.e.,
\begin{align*}
  \qv(t+1)                      & = \qv(t) + \dot{\qv}_\mathrm{cmd}(t)  \Delta T \\
  \dot{\qv}(t+1)                & = \dot{\qv}_\mathrm{cmd}(t)                    \\
  \dot{\xv}_{\mathrm{ee}} (t+1) & = \Jm_\mathrm{ee}(\qv(t+1)) \dot{\qv}(t+1).
\end{align*}
This procedure is repeated until the simulated agent reaches the goal pose $\xv = \xgoal$.
Whenever the \gls{ee} or a control point of the predictive agent comes close to an obstacle, i.e., $\norm{\dv} \leq \drange$, $\nagent$ new agents with different parameters are created for exploring different avoidance directions around an obstacle.
All agents are evaluated after a defined number of sampling steps by a cost function, which can be adapted to the desired robot behavior and required task.
The parameters of the best agent are then used to control the robot.
In order to ensure reactive behavior of the planner, all agent calculations are computed in parallel to the control command of the real robot.
Details on the multi-agent framework and extensive simulation results can be found in \cite{BeckerLilMulHad2021}.
In contrast to our previous work, the additional control points on the robot lead to a significant increase of agents, especially in the case of many obstacles.
Furthermore, many combinations of magnetic field vectors for the different control points will lead to infeasible trajectories or even to collisions.
Consequently, a considerable amount of computing power is wasted.
To compensate for these disadvantages, we introduce the concept of \gls{icf} in the next section.
%
%
\section{Informed Circular Fields}
The idea of \gls{icf} is to leverage the strengths of global and local motion planning strategies by extracting useful information from a global pre-planner, which are then transferred and evaluated by the predictive agents to improve the reactive motion generation.
The generation of a motion command in our \gls{icf} framework can be split into four distinct phases.
First, a global configuration space motion planner is used to create a (coarse) joint trajectory. Then, the global information is extracted in form of the magnetic field vectors for all control points and obstacles as a basis for CF-based reactive motion planning.
Note that when we refer to the control points here and in the following, the \gls{ee} is also included if not stated otherwise.
Subsequently, multiple predictive agents are created with the extracted magnetic field vectors to process and evaluate the global motion plan. The parameters of the best agent are then transferred to the real robot and used to generate the reactive joint velocity commands.
The different phases are executed in parallel with different sampling rates.
%
\subsection{Global Trajectory Generation}
The main purpose of the global pre-planner is to infer estimates for feasible global trajectories from the current robot pose to a goal pose in configuration space, i.e., finding non-conflicting avoidance directions around the obstacles for all control points is more important than generating short and smooth paths.
Thus, we use a rather coarse discretization of the global planner for shorter planning cycles $T_\mathrm{global}$.
The global planner is continuously running and restarted after $T_\mathrm{global}$ seconds with updated obstacle and robot information.
When a successful trajectory is found during the planning time, the trajectory is passed to the global information extraction module.
We use the MoveIt! motion planning framework for global trajectory generation because it features a variety of sampling-based planners and supports point clouds as a format for obstacle representation \citep{ColemanSucChiCor2014}.
However, other configuration space planners can also be used and exchanged easily.
Note that the global motion planning is done in a static snapshot of the environment while the actual reactive creation of the control signal (\cref{eq:q_cmd}) is always done with the most current environment information.
%
\subsection{Global Information Extraction} \label{sec:global_mag_field}
Whenever the global pre-planner finds a joint trajectory, the global information extraction module transforms this joint trajectory into Cartesian trajectories for each control point. Then, the avoidance direction of each control point around each obstacle is extracted from those Cartesian trajectories and the corresponding magnetic field vectors are calculated.
For each control point $i$ and for each obstacle $j$ described by the points $\xv_{\mathrm{o}_{j,k}} \in \mathbb{O}_j \subset \mathbb{R}^{3 \times m_j}$, the following calculations are performed.

\begin{figure}
      \centering
      \includegraphics[width=\columnwidth]{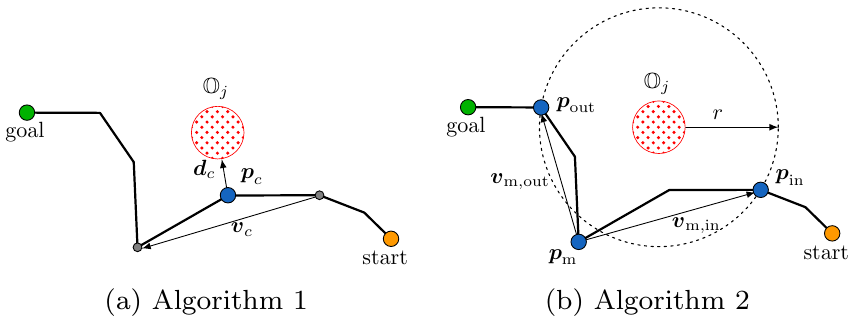}
      \vspace{-2ex}
      \caption{Visualization of global information extraction.}
      \label{fig:mag_field_extraction}
\end{figure}

\emph{Algorithm 1}
\begin{enumerate}
      \item Find the closest point $\pv_\mathrm{c}=\xv_\mathrm{cp_i}(\tau)$ of trajectory $\xv_\mathrm{cp_i}$ to obstacle $j$, where $\scalemath{0.85}{\tau = \underset{t\geq 0}{\arg\min}(\underset{\xv_{\mathrm{o}_{j,k}} \in \mathbb{O}_j}{\min} \norm{\xv_\mathrm{cp_i}(t) - \xv_{\mathrm{o}_{j,k}}})}$.
      \item Calculate the (approximate) direction of movement $\vv_\mathrm{c}=\xv_\mathrm{cp_i}(\tau+1) - \xv_\mathrm{cp_i}(\tau-1)$ at the closest point $\pv_\mathrm{c}$.
      \item Calculate $\dv_\mathrm{c}=\xv_{\mathrm{o}_{j,\mathrm{c}}}-\pv_\mathrm{c}$ pointing from $\pv_\mathrm{c}$ to the closest obstacle point $\xv_{\mathrm{o}_{j,\mathrm{c}}} = \underset{\xv_{\mathrm{o}_{j,k}}}{\arg\min}(\norm{\xv_{\mathrm{o}_{j,k}} - \pv_\mathrm{c}})$.
      \item Calculate the magnetic field vector $\bv_{i,j}= \frac{\dv_\mathrm{c} \times \vv_\mathrm{c}}{\norm{\dv_\mathrm{c} \times \vv_\mathrm{c}}}$.
\end{enumerate}
\emph{Algorithm 2}
\begin{enumerate}
      \item Find the first point $\pin=\xv_\mathrm{cp_i}(\tau_{\min})$ of the trajectory $\xv_\mathrm{cp_i}$ in a ball of radius $r$ around obstacle $j$, where $\tau_{\min} = \underset{t \geq 0, \xv_{\mathrm{o}_{j,k}} \in \mathbb{O}_j}{\min}\left( t \mid \norm{\xv_\mathrm{cp_i}(t) - \xv_{\mathrm{o}_{j,k}}} \leq r  \right)$.
      \item Find the last point $\pout=\xv_\mathrm{cp_i}(\tau_{\max})$ of the trajectory $\xv_\mathrm{cp_i}$ in a ball of radius $r$ around obstacle $j$, where $\tau_{\max} = \underset{t \geq 0}{\max} \underset{\xv_{\mathrm{o}_{j,k}} \in \mathbb{O}_j}{\min} \left( t \mid \norm{\xv_\mathrm{cp_i}(t) - \xv_{\mathrm{o}_{j,k}}} \leq r \right) $.
      \item Find the point $\pmid=\xv_\mathrm{cp_i}(\tau)$ with $\tau$ being the largest integer less than or equal to the midpoint in $[\tau_{\min},\tau_{\max}]$, i.e.,
            $\tau = \lfloor 0.5(\tau_{\max}+\tau_{\min})\rfloor$.
      \item Calculate the vectors $\vin = \pin - \pmid$ pointing from $\pmid$ to $\pin$ and $\vout = \pout - \pmid$ from $\pmid$ to $\pout$.
      \item Calculate the magnetic field vector \\$\bv_{i,j} = \frac{\vin \times \vout}{\norm{\vin \times \vout}}$.
\end{enumerate}
Visualizations of both algorithms are shown in \cref{fig:mag_field_extraction} in 2D example setups.
The quality of the resulting avoidance motion from both methods highly depends on the environment and the global trajectory.
While the first algorithm captures the avoidance motion only at the closest obstacle point, the second algorithm is used to recreate the resulting avoidance direction in a larger area around the obstacle.
Therefore, we use both methods to calculate magnetic field vectors, which are adopted by the predictive agents as described in the next section.
Nevertheless, we simplify a potentially complex avoidance motion to a single vector per obstacle and control point. Thus, the resulting motion of the \gls{icf} planner is expected to deviate from the global trajectory.
However, we would like to emphasize again that we only use the global planner to estimate possible, non-contradictory avoidance directions, which are subsequently simulated and evaluated before being used to enable reactive control of the real robot.
%
%
\subsection{Adapting the Multi-Agent Framework}
In order to adequately account for the new information from the global pre-planner, the process for creating and deleting predictive agents (compare \cref{sec:multi_agents}) is redesigned.
Instead of creating new agents when an obstacle is encountered, new agents are created when new magnetic field vectors are received.
In particular, one agent with the current best parameter set $\mathbb{P}_\mathrm{best}$ and the current best magnetic field vectors $\mathbb{B}_\mathrm{best}$ and two agents with the current best parameter set $\mathbb{P}_\mathrm{best}$ and the new magnetic field vector sets $\mathbb{B}_\mathrm{A1}$, $\mathbb{B}_\mathrm{A2}$ from Alg.~1 and Alg.~2 are created.
Afterwards, more agents can be created to simulate the robot motion with different parameters and the three magnetic field vector sets.
In this paper, new agents are created which use additional forces in the nullspace to push the robot into configurations with higher manipulability (details are omitted due to space limitations) but also simple adaptions of, e.g., the scaling factors could be simulated.
Note that an agent with the best parameter set but older environment data could still exist when new agents are created. However, the planning time of the global planner is typically much higher than the simulation of the agents and all agents finish before new magnetic field vectors are received.
Moreover, the trajectory of the real robot is continuously compared to the predicted trajectories of the agents. Whenever the deviation of a trajectory point reaches a defined limit, the respective agent is deleted.
Additionally, agent predictions are stopped and evaluated after a defined number of prediction steps $n_\mathrm{ps}$ and appended to the end of the waiting queue, preventing agents that take long suboptimal trajectories from blocking the available computation slots.
%
\subsection{Safety-Improving Fallback}
During motion planning, it may happen that none of the predicted trajectories match the current state of the environment, either because of unpredictable movements of the obstacles or because the robot cannot follow the predicted trajectories with sufficient accuracy. In this case, collision avoidance has absolute priority and the robot switches to a safety-improving fallback behavior. Here, the \gls{cf} forces on the control points on the robot structure (not the \gls{ee}) are replaced by repulsive forces with the scaling from \cite{LuoKoChuCha2014} in the form $\fcprep = -0.5\left( 1 + \tanh(\alpha - \beta \cdot \norm{\dv})\right) \ubar{\dv}$.
The safety-improving fallback is deactivated as soon as an agent finds a feasible trajectory to the goal pose.
%
%
\section{Evaluation}
\begin{figure*}%
  \centering
  \includegraphics[width=0.9\textwidth]{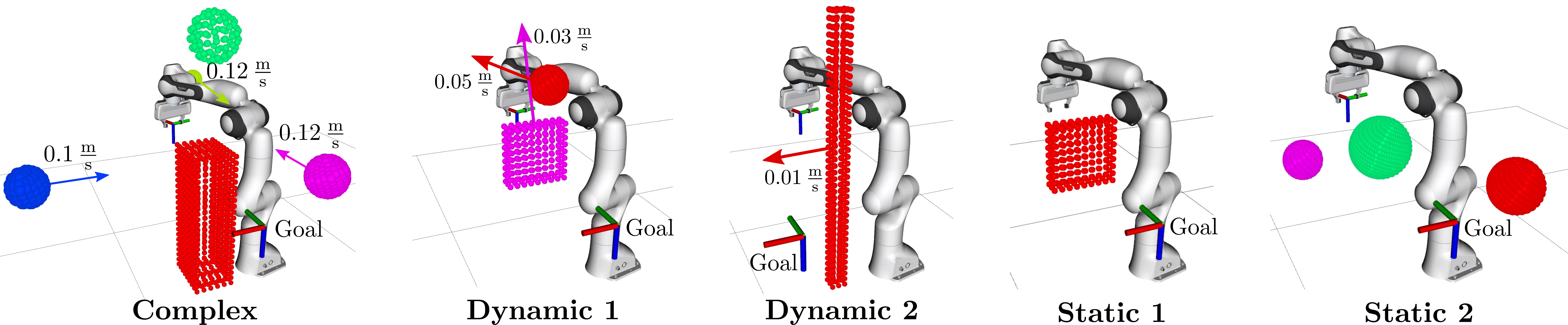}%
  \caption{Simulation environments used for comparisons.}%
  \label{fig:envs}%
\end{figure*}

In this section, we evaluate the performance of the \gls{icf} planner in five static and dynamic environments against other reactive and global motion planning approaches.
The evaluations are performed in a kinematic simulation without disturbances and with complete knowledge of the current positions and velocities of the obstacles.
We use a C++ implementation on a PC with an AMD Ryzen 9 5950X CPU with 16 cores, \SI{3.4}{GHz}.
Our simulations include complex environments with multiple dynamic obstacles which are much easier to follow in videos than images.
Thus, in addition to presenting the scenarios in \cref{fig:envs}, we uploaded videos of all simulations here: \url{https://doi.org/10.25835/erbbvx5h}.
%
For the comparison, we included two different empirically chosen pre-planners within our \gls{icf} framework, the \gls{prm} and the \gls{rrt}* planner.
Among the reactive control approaches, we use the original \gls{apf} approach from \cite{Khatib1986}, the \gls{cf} approach from \cite{AtakaLamAlt2018} and our extension of the \gls{cfp} from \cite{BeckerLilMulHad2021} introduced in \cref{sec:cfp_full_body}.
Note that the safety-improving fallback and our attractive force is also implemented for the locally reactive controllers to allow a fair comparison of the force command generation.
The global planners include the same sampling-based planners in their original version as used for our pre-planner, the \gls{prm} and \gls{rrt}* as well as the \gls{rrt}Connect and the optimizing STOMP planner from \cite{KalakrishnanChiThePas2011} due to their high prevalence and availability in the MoveIt! framework.
The sampling-based planners and the STOMP planner only consider static obstacles and are therefore only tested in the two static environments, in which a maximum planning time of \SI{0.5}{\second} was allowed. If no solution was found in this time, the attempt was considered a failure.
All non-deterministic planners were executed until ten successful runs were recorded for each planner in each environment.
Note that iteration time refers to the planning time for global planners (depicted with a * in \cref{tab:comp_ppa}) while it is used to specify the calculation time for a control command for the reactive planners.
Also note that the path duration depends highly on the allowed maximum velocities. While the \gls{ee} velocity of the force-based methods can be specified via a Cartesian maximum velocity, the random-based planners used here operate exclusively in joint space. Specifying an \gls{ee} velocity by means of constraints in the joint space is not straightforward, therefore, a measurement of the duration for the affected planners was omitted.\\
Overall, the results in \cref{tab:comp_ppa} show that the \gls{icf} planner is able to outperform the other tested motion planners.
Although its path length in static environments is (slightly) larger than the result from the optimizing STOMP planner, it still remains in a comparable range while additionally being able to reactively avoid dynamic obstacles.
The advantage of the \gls{icf} planner is then particularly noticeable in the second static environment. Even though the environment does not look complicated at first, a relatively complex joint movement of the robot is required in order not to collide with the obstacles, which leads to problems of the optimizing STOMP and \gls{rrt}*, reflected in the significantly lower success rates.
The \gls{cfp} planner also showed problems in this environment because of the extensive number of predictive agents that were created, also noticeable in the significantly increased iteration times. Therefore, no suitable combination of avoidance directions could be found and opposing forces on the control points led to a collision.
The complex environment requires simultaneous avoidance motions of the \gls{ee} and the body and was meant to test the limits of the motion planners; our \gls{icf} planner was the only planner that was able to successfully reach the goal without a collision. However, as can be seen in \cref{tab:comp_ppa}, it was not able to succeed in all runs. The \gls{icf} planner failed when the pre-planner was repeatedly not able to find a path to the goal in the given planning time of \SI{0.2}{\second}. Then, the \gls{icf} planner used the safety-improving fallback method and the obstacles pushed the robot into configurations which were close to the joint limits, where no avoidance was possible and a collision became inevitable.
Similarly, in one execution in the environment \emph{Dynamic 1}, the ICF-RRT* planner ended in a configuration close to the joint limits and was not able to reach the goal afterwards.
\begin{table}
  \centering
  \caption{Motion planner comparison.}\vspace{-1ex}
  \resizebox{0.95\columnwidth}{!}{%
    \begin{tabular}{llllcc} \toprule
      {Env.}                     & {Method}   & {Length [m]}  & {Duration [s]} & {Success}              & {Iter. Time [ms]} \\ \midrule

      \multirow{10}{*}{Static 1} & PRM        & 1.81          & -              & 100 \%                 & 190.0*            \\
                                 & RRTConnect & 1.51          & -              & 100 \%                 & 74.6*             \\
                                 & RRT*       & 1.71          & -              & 100 \%                 & 500.0*            \\
                                 & STOMP      & \textbf{1.34} & -              & 100 \%                 & 120.1*            \\
                                 & Khatib     & 1.65          & 17.81          & True                   & 0.020             \\
                                 & Ataka      & 1.86          & 19.56          & True                   & \textbf{0.018}    \\
                                 & \gls{cfp}  & 1.49          & 13.00          & True                   & 0.019             \\
                                 & ICF-PRM    & 1.41          & 11.88          & 100 \%                 & \textbf{0.018}    \\
                                 & ICF-RRT*   & 1.49          & \textbf{11.78} & 100 \%                 & 0.019             \\ \midrule

      \multirow{10}{*}{Static 2} & PRM        & 2.02          & -              & 100 \%                 & 237.3*            \\
                                 & RRTConnect & 2.01          & -              & 100 \%                 & 257.8*            \\
                                 & RRT*       & 2.00          & -              & 62 \%                  & 500.0*            \\
                                 & STOMP      & \textbf{1.72} & -              & 43 \%                  & 163.3*            \\
                                 & Khatib     & -             & -              & \textcolor{red}{False} & \textbf{0.039}    \\
                                 & Ataka      & -             & 60.00          & False                  & 0.045             \\
                                 & \gls{cfp}  & -             & -              & \textcolor{red}{False} & 0.209             \\
                                 & ICF-PRM    & 1.88          & 30.45          & 100 \%                 & 0.057             \\
                                 & ICF-RRT*   & 2.15          & \textbf{26.51} & 100 \%                 & 0.053             \\ \midrule

      \multirow{6}{*}{Dyn. 1}    & Khatib     & 1.50          & 11.46          & True                   & \textbf{0.029}    \\
                                 & Ataka      & 1.53          & 10.87          & True                   & 0.030             \\
                                 & \gls{cfp}  & -             & -              & \textcolor{red}{False} & 0.053             \\
                                 & ICF-PRM    & \textbf{1.31} & \textbf{12.55} & 100 \%                 & 0.035             \\
                                 & ICF-RRT*   & 1.48          & 13.13          & 91 \%                  & 0.033             \\ \midrule

      \multirow{6}{*}{Dyn. 2}    & Khatib     & 1.68          & 33.81          & True                   & 0.020             \\
                                 & Ataka      & -             & 60.00          & False                  & 0.021             \\
                                 & \gls{cfp}  & \textbf{1.47} & 12.75          & True                   & 0.029             \\
                                 & ICF-PRM    & 1.51          & \textbf{11.96} & 100 \%                 & 0.016             \\
                                 & ICF-RRT*   & 1.54          & 12.55          & 100 \%                 & \textbf{0.014}    \\ \midrule

      \multirow{6}{*}{Complex}   & Khatib     & -             & -              & \textcolor{red}{False} & 0.082             \\
                                 & Ataka      & -             & -              & \textcolor{red}{False} & 0.080             \\
                                 & \gls{cfp}  & -             & -              & \textcolor{red}{False} & 0.126             \\
                                 & ICF-PRM    & \textbf{2.12} & \textbf{27.42} & \textcolor{red}{71 \%} & \textbf{0.048}    \\
                                 & ICF-RRT*   & 2.20          & 28.46          & \textcolor{red}{62 \%} & 0.058             \\ \bottomrule
    \end{tabular}%
  }
  \label{tab:comp_ppa}
\end{table}

%
%
\section{Conclusion}
In this paper, the \gls{cfp} motion planner from \cite{BeckerLilMulHad2021} was applied to robotic manipulators and extended to the \gls{icf} algorithm by extracting information about feasible avoidance directions from a global pre-planner.
The \gls{icf} planner was tested in several simulations, where it performed well even in complex environments with multiple dynamic obstacles and demonstrated its advantage over other widespread global and locally reactive motion planners.
Future research includes improving the global information extraction module, which is currently not able to extract sufficient information when more involved avoidance motions of complex obstacles are necessary. Additionally, we are working on an implementation on a real robot.
%
%
\bibliography{IEEEabrv,ifacconf}             

\end{document}